\documentclass{article}

\PassOptionsToPackage{numbers,compress,sort}{natbib}
\usepackage[preprint]{nips_2018}

\usepackage[utf8]{inputenc} % allow utf-8 input
\usepackage[T1]{fontenc}    % use 8-bit T1 fonts
\usepackage{hyperref}       % hyperlinks
\usepackage{url}            % simple URL typesetting
\usepackage{booktabs}       % professional-quality tables
\usepackage{amsfonts}       % blackboard math symbols
\usepackage{nicefrac}       % compact symbols for 1/2, etc.
\usepackage{microtype}      % microtypography
\usepackage{dsfont}      % microtypography
\usepackage[algoruled,vlined]{algorithm2e}
\usepackage{subcaption,wrapfig,floatrow,amsmath}

\usepackage{tikz}
\usetikzlibrary{shapes,arrows.meta,calc,decorations.pathreplacing}

\title{Continual Learning via Neural Pruning}

\author{Siavash Golkar \\
	New York University\\
	\texttt{golkar@nyu.edu} \\
	\And 
	Michael Kagan\\
	SLAC National Accelerator Laboratory\\
	\texttt{makagan@slac.stanford.edu}
	\And 
	Kyunghyun Cho\\
	New York University\\
	Facebook AI Research \\
	CIFAR Azrieli Global Scholar \\
	\texttt{kyunghyun.cho@nyu.edu}
}

\begin{document}

\maketitle

\begin{abstract}
    We introduce Continual Learning via Neural Pruning~(CLNP), a new method aimed at lifelong learning in fixed capacity models based on neuronal model sparsification. In this method, subsequent tasks are trained using the inactive neurons and filters of the sparsified network and cause zero deterioration to the performance of previous tasks. In order to deal with the possible compromise between model sparsity and performance, we formalize and incorporate the concept of graceful forgetting: the idea that it is preferable to suffer a small amount of forgetting in a controlled manner if it helps regain network capacity and prevents uncontrolled loss of performance during the training of future tasks. CLNP also provides simple continual learning diagnostic tools in terms of the number of free neurons left for the training of future tasks as well as the number of neurons that are being reused. In particular,  we see in experiments that CLNP verifies and automatically takes advantage of the fact that the features of earlier layers are more transferable.   We show empirically that CLNP leads to significantly improved results over current weight elasticity based methods.
\end{abstract}

%%%%%%%%%%%%%%%%%%%%%%%%%%%%%%%%%%%%%%%%%%%%%%%%%%%%%%%%%%%%%%%%%%%%%%%%%%%%%%%%%%
\section{Introduction} \label{sec:intro}
%%%%%%%%%%%%%%%%%%%%%%%%%%%%%%%%%%%%%%%%%%%%%%%%%%%%%%%%%%%%%%%%%%%%%%%%%%%%%%%%%%

Continual learning, the ability of models to learn to solve new tasks beyond what has previously been trained, has garnered much attention from the machine learning community in recent years. This is driven in part by the practical advantages promised by continual learning schemes such as improved performance on subsequent tasks as well as a more efficient use of resources in machines with memory constraints. There is also great interest in continual learning from a more long term perspective, in that any approach towards artificial general intelligence needs to be able to continuously build on top of prior experiences.

As it stands today, the main obstacle in the path of effective continual learning is the problem of catastrophic forgetting: machines trained on new problems forget about the tasks that they were previously trained on. There are multiple approaches in the literature which seek to alleviate this problem, ranging from employing networks with many submodules~\cite{progressive,pathnet,dynamical_expnasion} to constraining the weights of the network that are deemed important for previous tasks~\cite{consolidation,synaptic_intelligence,power_law}. These approaches either require specialized genetic algorithm training schemes or still suffer catastrophic forgetting, albeit at a smaller rate. In particular, to the best of our knowledge, there is no form of guarantee regarding the performance of previous tasks among the fixed capacity models which use standard SGD training schemes.

In this work we introduce a simple fixed capacity continual learning scheme which can be trained using standard gradient descent methods and by construction suffers \emph{zero} deterioration on previously learned problems during the training of new tasks. In short, we take advantage of the over-parametrization of neural networks by using an activation based neural pruning sparsification scheme to train models which only use a fraction of their width. We then train subsequent tasks utilizing the unused capacity of the model. By cutting off certain connection in the network, we make sure that new tasks can take advantage of previously learned features but cause no interference in the pathways of the previously learned tasks.

\subsection*{Main contributions} 
\begin{itemize}
\item We introduce Continual Learning via Neural Pruning (CLNP), a simple and intuitive lifelong learning method with the following properties:
\begin{itemize}
    \item Given a network with activation based neuron sparsity trained on some previous tasks, CLNP trains new tasks utilizing the unused weights of the network in a manner which takes advantage of the features learned by the previous tasks while causing \emph{zero} catastrophic forgetting.
    \item CLNP provides simple diagnostics in the form of the number of remaining and reused neurons and filters when training each task. In particular, in experiments we see that CLNP verifies and automatically takes advantage of the fact that the features of earlier layers are more transferable.
\end{itemize}
\item We expand on the idea of graceful forgetting, the notion that it is preferable to suffer some small amount of forgetting in a controlled manner if it helps regain network capacity and prevents uncontrolled loss of performance during the training of future tasks. We use this idea to control the compromise between network sparsity and model accuracy in our approach.
\item We show empirically that using an activation based neural pruning sparsification scheme, we significantly outperform previous approaches based on weight elasticity on a number of benchmarks. We also demonstrate in one example that using a slightly more advanced variation of our sparsification method, the network suffers virtually no loss of performance, either from catastrophic forgetting or from sparsification.
\end{itemize}

The remainder of the paper is organized as follows. In Sec.~\ref{sec:method}, we provide the methodology of our approach. We first discuss the generalities of training non-destructively in sparsified regions of the network in Sec.~\ref{sec:general}. We then discuss our sparsification scheme as well as the idea of graceful forgetting as a compromise between sparsity and model performance in Sec.~\ref{sec:sparsification}. We provide an empirical test of our methodology in Sec.~\ref{sec:experiments}, first on ten tasks derived from the MNIST dataset in Sec.~\ref{sec:permnist} and then on a problem derived from CIFAR-10 and CIFAR-100 in Sec.~\ref{sec:icifar}. In both cases we show significant improvement over comparable prior work.

\section*{Related work}

\paragraph{Lifelong learning.} Previous work addressing the problem of catastrophic forgetting generally fall under two categories. In the first category, the model is comprised of many individual modules at each layer and forgetting is prevented either by routing the data through different modules~\cite{pathnet} or by successively adding new modules for each new task~\cite{progressive,dynamical_expnasion}. This approach often (but not always) has the advantage of suffering zero forgetting, however, the structure of these networks is specialized. In the case of~\cite{progressive,dynamical_expnasion}, the model is not fixed capacity and in the case of~\cite{pathnet} training is done using a tournament selection genetic algorithm. In the second category of approaches to lifelong learning the structure of the network as well as the training scheme are standard, and forgetting is addressed by constraining important weights from changing~\cite{consolidation,synaptic_intelligence,power_law}. These approaches, generally referred to as weight elasticity methods, have the advantage of having simpler training schemes but still suffer catastrophic forgetting, albeit at a smaller rate than unconstrained training.

The approach we take in this paper falls under the second category where the network structure and training scheme are simple and forgetting is prevented by constraining certain weights from changing. However, it shares the main advantage of the first category approaches in that we suffer zero catastrophic forgetting during the training of subsequent tasks. In particular, our method can be thought of as a second category adaptation and simplification of the path based approach of~\cite{pathnet} using activation based sparsification and the idea of graceful forgetting. Since our method is directly comparable to other second category approaches, we will provide quantitative comparisons with other methods in this category.

\paragraph{Network superposition.} The method we put forward in this paper can be thought of as a non-destructive way of superposing multiple instances of the same architecture using sparsification. There are previous works in this direction~\cite{fabrics,MixofExp}, however these are more along the lines of neural architecture search and are not aimed at preventing forgetting. The most relevant work in this direction is~\cite{superposition}, which can be thought of as a Fourier space implementation of our approach. However, in comparison to~\cite{superposition} where the superposition is approximate, our approach is exact and we have precise control over the relationship between the different models being superposed in terms of transfer learning. We will provide quantitative comparisons in our experiments. 

\paragraph{Sparsification.}
Sparsification of neural networks has a long history~\cite{brain_damage,sparsification,filter_pruning}. While sparsification is a crucial tool that we use, it is not in itself a focus of this work. For accessibility, we use a simple neuron/filter based sparsification scheme which can be thought of as a single iteration variation of~\cite{trimming} without fine-tuning. For a recent review of sparsification methods see~\cite{sparsification_review}.

Parameter based sparsity  (sparsity in the weight matrices) was used in~\cite{dynamical_expnasion} as a tool to reduce (but not eliminate) catastrophic forgetting and increase efficiency in a dynamically growing network. In contrast, our method is centered around activation based sparsity (the sparsity in the number of used filters/neurons), suffers zero catastrophic forgetting and is fixed capacity. Our method also bears some resemblance to~\cite{sharpening}, a forbear of clustering based methods put forward in the 90s.

%%%%%%%%%%%%%%%%%%%%%%%%%%%%%%%%%%%%%%%%%%%%%%%%%%%%%%%%%%%%%%%%%%%%%%%%%%%%%%%%%%
\section{Methodology} \label{sec:method}
%%%%%%%%%%%%%%%%%%%%%%%%%%%%%%%%%%%%%%%%%%%%%%%%%%%%%%%%%%%%%%%%%%%%%%%%%%%%%%%%%%

The core idea of our method is to take advantage of the fact that neural networks are vastly over-parametrized~\cite{overparametrization}. A manifestation of this over-parametrization is through the practice of sparsification, i.e. the compression of neural network with relatively little loss of performance~\cite{brain_damage,sparsification,sparsification_review}. As an example,~\citet{filter_pruning} show that VGG-16 can be compressed by more than 16 times, leaving more than 90\% of the connections unused. In this section we first show that given an activation based sparse network, we can leverage the unused capacity of the model to develop a continual learning scheme which suffers no catastrophic forgetting. We then discuss the idea of graceful forgetting to address the tension between sparsification and model performance in the context of lifelong learning.

It is important to differentiate activation based neuronal sparsity from parameter based weight sparsity. The former implies only a subset of the neurons or filters of each layer are active whereas the latter means that many of the weights are zero but all neurons are assumed to be active at all layers. In the remainder of the paper, when we mention sparsity, we are referring to activation based neuronal sparsity.

In what follows we will discuss sparsity for fully connected layers by looking at the individual neurons. The same argument goes through identically for individual channels of convolutional layers.

\subsection{Generalities}\label{sec:general}

Let us assume that we have a trained network which is sparse in the sense that only a subset of the neurons of the network are active. In effect, networks with this form of sparsity can be thought of as narrower networks embedded inside the original structure. There are many approaches that aim to train such sparse networks with little loss of performance (see for example Refs.~\cite{filter_pruning,trimming}). We will discuss our particular sparsification method in detail in Sec.~\ref{sec:sparsification}. 

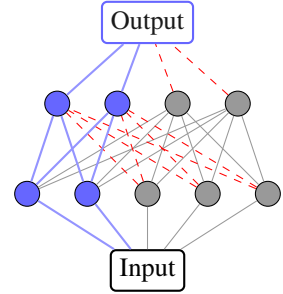
\begin{wrapfigure}{R}{0.35\textwidth}
% \vspace{-5pt}
\centering
\begin{tikzpicture}[darkstyle/.style={circle,draw,fill=gray!40,minimum size=20}]
\definecolor{c3}{rgb}{0.6,0.6,0.6}
\definecolor{c1}{rgb}{0.4,0.4,1}
\pgfmathsetmacro{\xl}{0.8}
\pgfmathsetmacro{\yl}{1.2}
\pgfmathsetmacro{\actf}{2}
\pgfmathsetmacro{\acts}{2}
\pgfmathsetmacro{\inactf}{3}
\pgfmathsetmacro{\inacts}{2}
\node [rounded corners=2pt,draw,rectangle,thick] (input) at (0,-0.85*\yl) {Input};
\node [rounded corners=2pt,draw,rectangle,color=c1,thick,text=black!90] (output) at (0,1.9*\yl) {Output};

    \foreach \y/\i/\a in {0/\inactf/\actf,1/\inacts/\acts}
        {\foreach \x in {1,...,\a}
            \node [draw,fill=c1,ellipse]  (act\x\y) at (-0.5*\xl-0.5*\i*\xl-0.5*\a*\xl+\xl*\x,\yl*\y) {};
        \foreach \x in {1,...,\i}
            \node [draw,fill=c3,ellipse]  (inact\x\y) at (-0.5*\xl-0.5*\i*\xl+0.5*\a*\xl+\xl*\x,\yl*\y) {};
        }

    % Drawing the free weights
    \foreach \xb in {1,...,\inacts}
        {\foreach \xa in {1,...,\actf}
            \draw [color=c3] (act\xa0)--(inact\xb1) ;
        \foreach \xa in {1,...,\inactf}
            \draw [color=c3] (inact\xa0)--(inact\xb1) ;}
    
    % Drawing the active and interference weights
    \foreach \xb in {1,...,\acts}
        {\foreach \xa in {1,...,\actf}
            \draw [color=c1!70,thick] (act\xa0)--(act\xb1) ;
        \foreach \xa in {1,...,\inactf}
            \draw [color=red,dashed] (inact\xa0)--(act\xb1) ;}
            
    % Drawing the input/output nodes
    \foreach \endpoint/\limit/\layer in {input/\actf/0,output/\acts/1}
        \foreach \x in {1,...,\limit}
            \draw [color=c1!70,thick] (act\x\layer)--(\endpoint) ;
            
    \foreach \x in {1,...,\inactf}
        \draw [color=c3] (inact\x0)--(input) ;
    \foreach \x in {1,...,\inacts}
        \draw [color=red,dashed] (inact\x1)--(output) ;
        
\end{tikzpicture}
\caption{\small The partition of an network with neuronal sparsity into active, inactive and interference parts.}\label{fig:split}
\end{wrapfigure}

Fig.~\ref{fig:split} shows a cartoon of a network with activation based neuronal sparsity, where the active and inactive neurons are respectively denoted by blue and grey nodes. Based on the connectivity structure, the weights of the network can also be split into three classes. First we have the active weights $W^{\textrm{\tiny act}}$ which connect active nodes to active nodes. These are denoted in blue in Fig.~\ref{fig:split}. Next we have the weights which connect any node to inactive nodes, we call these the free weights $W^{\textrm{\tiny free}}$, denoted in grey in Fig.~\ref{fig:split}. Finally we have the weights which connect the inactive nodes to the active nodes, we call these the interference weights $W^{\textrm{\tiny int}}$, denoted in red dashed lines in Fig.~\ref{fig:split}. A more precise definition of the active and inactive neurons and weights is given in Sec.~\ref{sec:sparsification}.

The crux of our approach is the simple observation that if all the interference weights $W^{\textrm{\tiny int}}$ are set to zero, the free weights $W^{\textrm{\tiny free}}$ can be changed arbitrarily without causing any change whatsoever to the output of the network. We can therefore utilize these weights to train new tasks without causing any harm to the performance of the previous tasks. 

Note that we can further split the free weights into two groups. First, the weights which connect active nodes to inactive nodes. These are the weights that take advantage of previously learned features and are therefore responsible for transfer learning throughout the network. We also have the weights that connect inactive nodes to inactive nodes. These weights can form completely new pathways to the input and train new features. A complimentary diagnostic for the amount of transfer learning taking place is the number of new active neurons at each layer after the training of subsequent tasks. Given that an efficient sparse training scheme would not need to relearn the features that are already present in the network, the number of new neurons grown at each stage of training is an indicator of the sufficiency of the already learned features for the purposes of the new task.  We will see more of this point in Sec.~\ref{sec:experiments}.

\begin{figure}[ht]
\centering
\begin{tikzpicture}[darkstyle/.style={circle,draw,fill=gray!40,minimum size=20}]
\definecolor{c3}{rgb}{0.6,0.6,0.6}
\definecolor{c2}{rgb}{0.1,0.85,0.1}
\definecolor{c1}{rgb}{0.4,0.4,1}

\pgfmathsetmacro{\yoffset}{-1.7}

\pgfmathsetmacro{\xl}{0.7}
\pgfmathsetmacro{\yl}{1.2}

\pgfmathsetmacro{\actf}{2}
\pgfmathsetmacro{\acts}{2}
\pgfmathsetmacro{\inactf}{3}
\pgfmathsetmacro{\inacts}{2}
\node [rounded corners=2pt,draw,rectangle,thick] (input) at (0,-0.85*\yl) {Input};
\node [rounded corners=2pt,draw,rectangle,color=c1,thick,text=black!90] (output) at (0,1.9*\yl) {Output};

    \foreach \y/\i/\a in {0/\inactf/\actf,1/\inacts/\acts}
        {\foreach \x in {1,...,\a}
            \node [draw,fill=c1,ellipse]  (act\x\y) at (-0.5*\xl-0.5*\i*\xl-0.5*\a*\xl+\xl*\x,\yl*\y) {};
        \foreach \x in {1,...,\i}
            \node [draw,fill=c3,ellipse]  (inact\x\y) at (-0.5*\xl-0.5*\i*\xl+0.5*\a*\xl+\xl*\x,\yl*\y) {};
        }

    % Drawing the free weights
    \foreach \xb in {1,...,\inacts}
        {\foreach \xa in {1,...,\actf}
            \draw [color=c3] (act\xa0)--(inact\xb1) ;
        \foreach \xa in {1,...,\inactf}
            \draw [color=c3] (inact\xa0)--(inact\xb1) ;}
    
    % Drawing the active 
    \foreach \xb in {1,...,\acts}
        {\foreach \xa in {1,...,\actf}
            \draw [color=c1!70,thick] (act\xa0)--(act\xb1) ;}
            
    % Drawing the input/output nodes
    \foreach \endpoint/\limit/\layer in {input/\actf/0,output/\acts/1}
        \foreach \x in {1,...,\limit}
            \draw [color=c1!70,thick] (act\x\layer)--(\endpoint) ;
            
    \foreach \x in {1,...,\inactf}
        \draw [color=c3] (inact\x0)--(input) ;
        
%%%%%%%%%%%%%%%%%%%%%%%%%%%%%%%%%%%%%%%%%%%%%%%%%%%%%%%%%%%%%%%%%%%%%%%%%%%%%%%%%%%%%%%%%%%%%%%%%%%%%%%%%%%%        
        
\pgfmathsetmacro{\mhxoffset}{-4.8}
\pgfmathsetmacro{\mhactonef}{2}
\pgfmathsetmacro{\mhactones}{2}
\pgfmathsetmacro{\mhacttwof}{2}
\pgfmathsetmacro{\mhacttwos}{1}
\pgfmathsetmacro{\mhinactf}{1}
\pgfmathsetmacro{\mhinacts}{1}
\node [rounded corners=2pt,draw,rectangle,thick] (mhinput) at (\mhxoffset,-0.85*\yl+\yoffset) {Input};
\node [rounded corners=2pt,draw,rectangle,color=c1,thick,text=black!90] (mhhead1) at (\mhxoffset-0.5*\xl-0.5*\mhinacts*\xl-0*\mhactones*\xl-0.5*\mhacttwos*\xl,1.9*\yl+\yoffset) {Head 1};
\node [rounded corners=2pt,draw,rectangle,color=c2,thick,text=black!90] (mhhead2) at (\mhxoffset+0.5*\xl-0.5*\mhinacts*\xl+0.5*\mhactones*\xl-0*\mhacttwos*\xl,1.9*\yl+\yoffset) {Head 2};

    \foreach \y/\i/\ao/\at in {0/\mhinactf/\mhactonef/\mhacttwof,1/\mhinacts/\mhactones/\mhacttwos}
        {\foreach \x in {1,...,\ao}
            \node [draw,fill=c1,ellipse]  (mhactone\x\y) at (\mhxoffset-0.5*\xl-0.5*\i*\xl-0.5*\at*\xl-0.5*\ao*\xl+\xl*\x,\yl*\y+\yoffset) {};
        \foreach \x in {1,...,\at}
            \node [draw,fill=c2,ellipse]  (mhacttwo\x\y) at (\mhxoffset-0.5*\xl-0.5*\i*\xl-0.5*\at*\xl+0.5*\ao*\xl+\xl*\x,\yl*\y+\yoffset) {};
        \foreach \x in {1,...,\i}
            \node [draw,fill=c3,ellipse]  (mhinact\x\y) at (\mhxoffset-0.5*\xl-0.5*\i*\xl+0.5*\ao*\xl+0.5*\at*\xl+\xl*\x,\yl*\y+\yoffset) {};
        }

    % Drawing the free weights
    \foreach \xb in {1,...,\mhinacts}
        {\foreach \xa in {1,...,\mhactonef}
            \draw [color=c3] (mhactone\xa0)--(mhinact\xb1) ;
        \foreach \xa in {1,...,\mhacttwof}
            \draw [color=c3] (mhacttwo\xa0)--(mhinact\xb1) ;
        \foreach \xa in {1,...,\mhinactf}
            \draw [color=c3] (mhinact\xa0)--(mhinact\xb1) ;}
    
    % Drawing the active 
    \foreach \xb in {1,...,\mhactones}
        {\foreach \xa in {1,...,\mhactonef}
            \draw [color=c1!70,thick] (mhactone\xa0)--(mhactone\xb1) ;}
    \foreach \xb in {1,...,\mhacttwos}
        {\foreach \xa in {1,...,\mhacttwof}
            \draw [color=c2!70,thick] (mhacttwo\xa0)--(mhacttwo\xb1) ;}
    \foreach \xb in {1,...,\mhactonef}      
        \foreach \xa in {1,...,\mhacttwos}
            \draw [color=c2!70,thick] (mhactone\xb0)--(mhacttwo\xa1) ;
            
    % Drawing the input
    \foreach \x in {1,...,\mhactonef}
        \draw [color=c1!70,thick] (mhactone\x0)--(mhinput) ;
    \foreach \x in {1,...,\mhacttwof}
        \draw [color=c2!70,thick] (mhacttwo\x0)--(mhinput) ;
    \foreach \x in {1,...,\mhinactf}
        \draw [color=c3] (mhinact\x0)--(mhinput) ;
        
    % Drawing the output
    \foreach \x in {1,...,\mhactones}
        {\draw [color=c1!70,thick] (mhactone\x1)--(mhhead1) ;
        \draw [color=c2!70,thick] (mhactone\x1)--(mhhead2) ;}
    \foreach \x in {1,...,\mhacttwos}
        \draw [color=c2!70,thick] (mhacttwo\x1)--(mhhead2) ;
%%%%%%%%%%%%%%%%%%%%%%%%%%%%%%%%%%%%%%%%%%%%%%%%%%%%%%%%%%%%%%%%%%%%%%%%%%%%%%%%%%%%%%%%%%%%%%%%%%%%%%%%%%%%%%%%%%%%%%%%%%

\pgfmathsetmacro{\shxoffset}{4.8}
\pgfmathsetmacro{\shactonef}{2}
\pgfmathsetmacro{\shactones}{2}
\pgfmathsetmacro{\shacttwof}{2}
\pgfmathsetmacro{\shacttwos}{1}
\pgfmathsetmacro{\shinactf}{1}
\pgfmathsetmacro{\shinacts}{1}
\node [rounded corners=2pt,draw,rectangle,thick] (shinput) at (\shxoffset,-0.85*\yl+\yoffset) {Input};
\node [rounded corners=2pt,draw,rectangle,color=c1,thick,text=black!90] (shoutput) at (\shxoffset,1.9*\yl+\yoffset) {Output};
\node [rounded corners=3pt,draw,rectangle,color=c2,thick,text=black!90, minimum width=1.37cm, minimum height = 0.72cm] (shoutput2) at (\shxoffset,1.9*\yl+\yoffset) {Output};

    \foreach \y/\i/\ao/\at in {0/\shinactf/\shactonef/\shacttwof,1/\shinacts/\shactones/\shacttwos}
        {\foreach \x in {1,...,\ao}
            \node [draw,fill=c1,ellipse]  (shactone\x\y) at (\shxoffset-0.5*\xl-0.5*\i*\xl-0.5*\at*\xl-0.5*\ao*\xl+\xl*\x,\yl*\y+\yoffset) {};
        \foreach \x in {1,...,\at}
            \node [draw,fill=c2,ellipse]  (shacttwo\x\y) at (\shxoffset-0.5*\xl-0.5*\i*\xl-0.5*\at*\xl+0.5*\ao*\xl+\xl*\x,\yl*\y+\yoffset) {};
        \foreach \x in {1,...,\i}
            \node [draw,fill=c3,ellipse]  (shinact\x\y) at (\shxoffset-0.5*\xl-0.5*\i*\xl+0.5*\ao*\xl+0.5*\at*\xl+\xl*\x,\yl*\y+\yoffset) {};
        }

    % Drawing the free weights
    \foreach \xb in {1,...,\shinacts}
        {\foreach \xa in {1,...,\shactonef}
            \draw [color=c3] (shactone\xa0)--(shinact\xb1) ;
        \foreach \xa in {1,...,\shacttwof}
            \draw [color=c3] (shacttwo\xa0)--(shinact\xb1) ;
        \foreach \xa in {1,...,\shinactf}
            \draw [color=c3] (shinact\xa0)--(shinact\xb1) ;}
    
    % Drawing the active 
    \foreach \xb in {1,...,\shactones}
        {\foreach \xa in {1,...,\shactonef}
            \draw [color=c1!70,thick] (shactone\xa0)--(shactone\xb1) ;}
    \foreach \xb in {1,...,\shacttwos}
        {\foreach \xa in {1,...,\shacttwof}
            \draw [color=c2!70,thick] (shacttwo\xa0)--(shacttwo\xb1) ;}
    \foreach \xb in {1,...,\shactonef}      
        \foreach \xa in {1,...,\shacttwos}
            \draw [color=c2!70,thick] (shactone\xb0)--(shacttwo\xa1) ;
            
    % Drawing the input
    \foreach \x in {1,...,\shactonef}
        \draw [color=c1!70,thick] (shactone\x0)--(shinput) ;
    \foreach \x in {1,...,\shacttwof}
        \draw [color=c2!70,thick] (shacttwo\x0)--(shinput) ;
    \foreach \x in {1,...,\shinactf}
        \draw [color=c3] (shinact\x0)--(shinput) ;
        
    % Drawing the output
    \foreach \x in {1,...,\shactones}
        \draw [color=c1!70,thick] (shactone\x1)--(shoutput) ;
    \foreach \x in {1,...,\shacttwos}
        \draw [color=c2!70,thick] (shacttwo\x1)--(shoutput2) ;
        
    % \draw [dashed, color=c1!70,thick] (\shxoffset+0.1*\xl,1.35*\yl) -- (\shxoffset+1.9*\xl,1.35*\yl);
    % \draw [dashed, color=c2!70,thick] (\shxoffset-1.8*\xl,1.28*\yl) -- (\shxoffset-0.15*\xl,1.28*\yl);
    % \draw [dashed, color=c2!70,thick] (\shxoffset+0.9*\xl,1.28*\yl) -- (\shxoffset+1.9*\xl,1.28*\yl);

    \draw [color=c1!70,thick] (\shxoffset+0*\xl,1.38*\yl+\yoffset) -- (\shxoffset+1.9*\xl,1.38*\yl+\yoffset);
    \draw [dotted,color=c1!70,thick] (\shxoffset-1.8*\xl,1.38*\yl+\yoffset) -- (\shxoffset+0*\xl,1.38*\yl+\yoffset);
    \draw [color=c2!70,thick] (\shxoffset-1.8*\xl,1.31*\yl+\yoffset) -- (\shxoffset-0*\xl,1.31*\yl+\yoffset);
    \draw [dotted,color=c2!70,thick] (\shxoffset-0*\xl,1.31*\yl+\yoffset) -- (\shxoffset+1.0*\xl,1.31*\yl+\yoffset);
    \draw [color=c2!70,thick] (\shxoffset+1.0*\xl,1.31*\yl+\yoffset) -- (\shxoffset+1.9*\xl,1.31*\yl+\yoffset);
    
    % \draw [color=c1!70,thick] (\shxoffset+0*\xl,1.31*\yl) -- (\shxoffset+1.9*\xl,1.31*\yl);
    % \draw [dotted,color=c1!70,thick] (\shxoffset-1.8*\xl,1.31*\yl) -- (\shxoffset+0*\xl,1.31*\yl);
    % \draw [color=c2!70,thick] (\shxoffset-1.8*\xl,1.38*\yl) -- (\shxoffset-0*\xl,1.38*\yl);
    % \draw [dotted,color=c2!70,thick] (\shxoffset-0*\xl,1.38*\yl) -- (\shxoffset+0.9*\xl,1.38*\yl);
    % \draw [color=c2!70,thick] (\shxoffset+0.9*\xl,1.38*\yl) -- (\shxoffset+1.9*\xl,1.38*\yl);
    
\draw [->,thick] (0.6*\mhxoffset*\xl,0.5*\yl+2.2*\yoffset/5)--(0.8*\mhxoffset*\xl,0.5*\yl+3.5*\yoffset/5);
\draw [->,thick] (0.6*\shxoffset*\xl,0.5*\yl+2.2*\yoffset/5)--(0.8*\shxoffset*\xl,0.5*\yl+3.5*\yoffset/5);

\end{tikzpicture}
\caption{\small The output structure of multi-task learning networks. The sparsified network trained on task 1 is presented in the center after having the interference weights severed. The multi-head and single-head expansion of this network trained on task 2 are presented on the left and right respectively. In both cases, the output of the model on task 1 remains unchanged during training task 2, i.e. the green connections do not affect the ouput of the blue sub-network. }\label{fig:heads}
\end{figure}
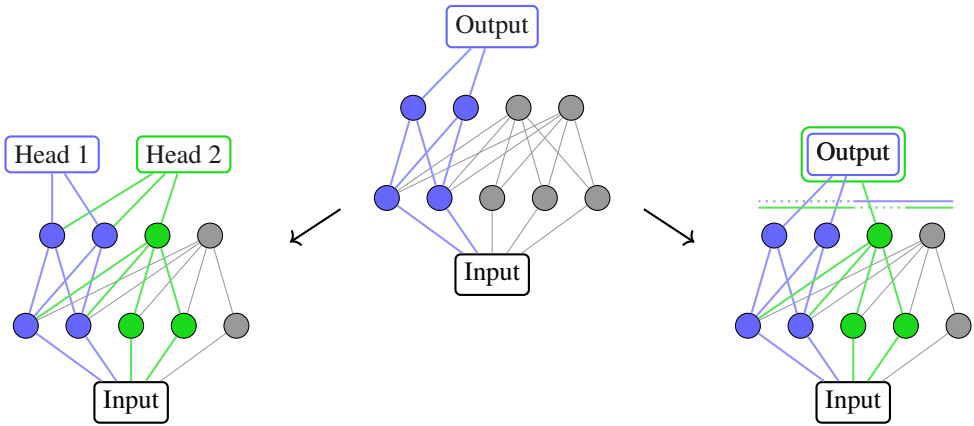

\paragraph{Output architecture.} In order to fully flesh out a continual learning scheme, we need to specify the connectivity structure of the output nodes. There are two intuitive routes that we can take. We demonstrate these in~Fig.~\ref{fig:heads}, where in the middle we have the same structure as in Fig.~\ref{fig:split}, but with the interference weights put to zero. In order to train a new task, one option is to use a new output layer (i.e. a new head) while saving the previous output layer. This option, demonstrated in Fig.~\ref{fig:heads} on the left, is known as the multi-head approach and is standard in continual learning. Because each new output layer comes with its own sets of weights which connect to the final hidden layer neurons, this method is not a fully fixed capacity method. Note that in our approach to continual learning, training a multi-head network with a fully depleted core structure, i.e. a network where are no more free neurons left, is equivalent to final layer transfer learning.

In scenarios where the output layer of the different tasks are structurally compatible, for example when all tasks are classification on the same number of classes, we can use a single-head approach. Demonstrated in Fig.~\ref{fig:heads} on the right, in this approach we use the same output layer for all tasks, but for each task, we mask out the neurons of the final hidden layer that were trained on other tasks. In the case of Fig.~\ref{fig:heads}, only green nodes in the final hidden layer are connected to the output for the second task and only blue nodes for the first task. This is equivalent to a dynamic partitioning of the final hidden layer into multiple  unequal sized parts, one part for each task. In practice this is done using a multiplicative masking operation with a task dependent mask, denoted in Fig.~\ref{fig:heads} by dashed lines after the final hidden layer. This structure is truly fixed capacity, but is more restrictive to train than its multi-head counterpart. Note that since an unconstrained single-head structure would cause a large amount of interference between tasks, until now it has not been a viable option for weight elasticity based approaches to continual learning.

\subsection{Methodology details}\label{sec:sparsification}

In what follows we will assume that the we are using Rectifier Linear Units (ReLU~\cite{relu}). While we have only tested our methodology with ReLU networks, we expect it to work similarly with other activations.

\paragraph{Sparsification.} So far in this section we have shown that given a sparse network trained on a number of tasks, we can train the network on new tasks without suffering any catastrophic forgetting. We now discuss the specific sparsification scheme that we use throughout this paper, which is similar in spirit to the network trimming approach put forward in Ref.~\cite{trimming}. 

Our sparsification method is comprised of two parts. First, during the training of each task, we add an $L^1$ weight regulator to promote sparsity in the network and to regulate the magnitude of the weights of the network. The coefficient of $\alpha$ of this regulator is a hyperparameter of our approach. Also, since different layers have different weight distributions, we can gain more control over the amount of sparsity in each layer by choosing a different $\alpha$ for each layer. The second part of our sparsification scheme is post-training neuron pruning based on the average activity of each neuron. Note that the most efficient sparsification algorithms include a third part which involves adjusting the surviving weights of the network after pruning. This step is referred to as fine-tuning and is done by retraining the network for a few epochs while only updating the weights which survive sparsification. This causes the model to regain some of its lost performance because of pruning. To achieve a yet higher level of sparsity, one can iterate the pruning and fine-tuning steps multiple times. In this paper, we only perform one iteration of pruning for simplicity. We also skip the fine tuning step, unless otherwise specified. 

In Sec.~\ref{sec:general}, we partitioned the network into active and inactive parts. A precise definition of these different partitions is as follows. Given network $N$, comprised of $L$ layers, we denote the neurons of each layer as $N_l$ with $l=1\cdots L$. Let us also assume that the network $N$ has been trained on dataset $S$. In order to find the active and inactive neurons of the network, we compute the average activity over the entire dataset $S$ for each individual neuron. We identify the active neurons $N_l^{\textrm{\tiny{act}}}$, i.e. the blue nodes in Fig.~\ref{fig:split}, as those whose average activation exceeds some threshold parameter $\theta$:  
\begin{equation}
N_l^{\textrm{\tiny{act}}} = \{ N_l \;|\; \mathds{E}_{S} \big(\,N_l\,\big) > \theta \}.\notag
% \label{eq:first_task_neurons}
\end{equation}
The inactive neurons are taken as the complement  $N_l^{\textrm{\tiny{inact}}} = N_l \setminus N_l^{\textrm{\tiny{act}}}$. The threshold value $\theta$ is a post-training hyperparameter of our approach. Similar to the $L^1$ weight regulator hyperparameter~$\alpha$, $\theta$ can take different values for the different layers. Furthermore, if $\theta=0$, $N_l^{\textrm{\tiny{inact}}}$ would be given by the neurons in the network which are completely dead and the function being computed by the network is entirely captured in $N_l^{\textrm{\tiny{act}}}$.  We can therefore view $N_l^{\textrm{\tiny{act}}}$ as a compression of the network into a sub-network of smaller width. Based on their connectivity structure, the weights of each layer are again divided into active, free and interference parts, respectively corresponding to the blue, grey and red lines in Fig.~\ref{fig:split}. The overall algorithm of our approach in its multi-head incarnation is given in Alg.~\ref{alg:multi_head} (Here, we use notation $W\big(A \to B\big)$ to denotes the subset of weights in $W$ which connect neurons $A$ to neurons $B$). 

\begin{algorithm}[ht]
\small
\SetKwFor{RepTimes}{repeat}{times}{end}
\BlankLine
\KwData{datasets $S = \{S_i\}$, network with $l$'th layer neurons $N_l$}

\BlankLine
$N_l^{\textrm{\tiny{act}}}, W^{\tiny\textrm{int}} \leftarrow \emptyset$\;
$W^{\tiny\textrm{free}} \longleftarrow W\big(N_l \to N_{l+1}\big)$\;
\For{$S_i \in S$}{
    $W^{\tiny\textrm{int}}\longleftarrow 0$\;
    Re-initialize $W^{\tiny\textrm{free}}$\;
    Place new head $N_L^{(i)}$\;
    Train on $S_i$ updating only $W^{\tiny\textrm{free}}$\;
    $N_l^{\textrm{act}} \longleftarrow N_l^{\textrm{act}} \cup \{ N_l \;|\; \mathds{E}_{S_i} \big(\,|N_l|\,\big) > \theta \}$\;
    $N_l^{\textrm{\tiny{inact}}} \longleftarrow N_l \setminus N_l^{(i)}$\;
    % $W^{(i)} \longleftarrow W\big(N_l^{(i)} \to N_{l+1}^{(i)}\big)$\;
    $W^{\tiny\textrm{int}} \longleftarrow W\big(N_l^{\tiny\textrm{inact}} \to N_{l+1}^{(1)}\big)$\;
    $W^{\tiny\textrm{free}} \longleftarrow W\big(N_l \to N_{l+1}^{\tiny\textrm{inact}}\big)$\;
}
\caption{Continual Learning via Neural Pruning (CLNP) - Multi-head}
\label{alg:multi_head}
\end{algorithm}

\paragraph{Graceful forgetting.} While sparsity is crucial in our approach for the training of later tasks, care needs to be taken so as not to overly sparsify and thereby reduce the model's performance. In practice, model sparsity has the same relationship with generalization as other regularization schemes. As sparsity increases, initially the generalization performance of the model improves. However, as we push our sparsity knobs (i.e. the $L^1$ regulator and activity threshold) higher and make the network sparser, eventually both training and validation accuracy will suffer and the network fails to fit the data properly. This means that in choosing these hyperparameters, we have to make a compromise between model performance and remaining network capacity for future tasks. 

This brings us to a subject which is often overlooked in lifelong learning literature generally referred to as graceful forgetting. This is the general notion that it would be preferable to sacrifice a tiny bit of accuracy in a controlled manner, if it reduces catastrophic forgetting of this task and helps in the training of future tasks. We believe any successful fixed capacity continual learning algorithm needs to implement some form of graceful forgetting scheme. In our approach, graceful forgetting is implemented through the sparsity vs. performance compromise. In other words, after the training of each task, we sparsify the model up to some acceptable level of performance loss in a controlled manner. We then move on to subsequent tasks knowing that the model no longer suffers any further deterioration from training future tasks. This has to be contrasted with other weight elasticity approaches which use soft constraints on the weights of the network and cannot guarantee future performance of previously trained tasks. We will see in the next section that using the exact same network structure, our approach leads to noticeably improved results over existing methods.

Explicitly, the choice of sparsity hyperparameters is made based on this idea of graceful forgetting as follows. As is standard practice,  we split the dataset for each task into training, validation and test sets. We scan over a range of hyperparameters (i.e. $\alpha$, the $L^1$ weight regulator and $\xi$, the learning rate)  using grid search and note the value of the best validation accuracy across all hyperparameters. We then pick the models which achieve validation accuracy within a margin of $m\%$ of this best validation accuracy. The margin parameter $m$ controls how much we are willing to compromise on accuracy to regain capacity and in experiments we take it to be generally in the range of $0.05\%$ to $2\%$ depending on the task. We sparsify the picked models using the highest activation threshold $\theta$ such that the model remains within this margin of the best validation accuracy. We finally pick the hyperparameters which give the highest sparsity among these models. In this way, we efficiently find the hyperparameters which afford the highest sparsity model with validation accuracy within $m\%$ of its highest value. 

% A cartoon of this method is given in Fig.~\ref{fig:sparsity_v_accuracy} where the result of each hyperparameter is denoted by a circle. The green nodes denote the models which are within the $m\%$ margin of the best validation accuracy and the dashed line represents the change in the model sparsity and validation accuracy as we increase $\theta$ from zero. The chosen hyperparameters are those which correspond to the model with the highest sparsity after this procedure (denoted by the square in Fig.~\ref{fig:sparsity_v_accuracy}).

After pruning away the unused weights and neurons of the model with the hyperparameters chosen as above, we report the test accuracy of the sparsified network. It should be noted that this algorithm for training and hyperparameter grid search does not incur any significant additional computational burden over standard practice. The hyperparameter search is performed in standard fashion, and the additional steps of selecting networks within the acceptable margin, scanning the threshold, and selecting the highest sparsity network only require evaluation and do not include any additional network training.

%%%%%%%%%%%%%%%%%%%%%%%%%%%%%%%%%%%%%%%%%%%%%%%%%%%%%%%%%%%%%%%%%%%%%%%%%%%%%%%%%%
\section{Experiments} \label{sec:experiments}
%%%%%%%%%%%%%%%%%%%%%%%%%%%%%%%%%%%%%%%%%%%%%%%%%%%%%%%%%%%%%%%%%%%%%%%%%%%%%%%%%%
We evaluated our approach for continual learning on the permuted MNIST~\cite{MNIST}, and split versions of CIFAR-10 and CIFAR-100~\cite{CIFAR} and compare to previous results.

\subsection{permuted MNIST} \label{sec:permnist}

In this experiment, we look at the performance of our approach on ten tasks derived from the MNIST dataset via ten random permutations of the pixels. To compare with previous work, we choose the same structure and hyperparameters as in Ref.~\cite{synaptic_intelligence}: an MLP with two hidden layers, each with 2000 neurons and ReLU activation and a softmax multi-class cross-entropy loss trained with Adam optimizer and batch size 256. We make a small modification to the structure of the network: as opposed to Refs.~\cite{synaptic_intelligence,superposition} which use a multi-head structure, we employ a single-head network. This makes our network a truly fixed capacity structure and renders the continual learning task more challenging. 

\setlength{\columnsep}{15pt}%
\setlength{\intextsep}{40pt}
\begin{wraptable}{R}{0.40\textwidth}
  \vspace{-10pt}
  \scalebox{0.75}{
  \begin{tabular}{cc}
    \toprule
    Method     &  Accuracy (\%) \\
    \midrule
    Single Task SGD                 &   $98.48 \pm 0.05$ \\
    \midrule
   ~\citet{consolidation}           &    97.0           \\
   ~\citet{synaptic_intelligence}   &    97.2            \\
   ~\citet{superposition}           &    97.6           \\
    % \midrule
    CLNP (ours)         &   $98.42\pm 0.04$  \\
    \bottomrule
    \end{tabular}}
  \caption{\small Comparison of test accuracy averaged over 10 permuted MNIST tasks.}
  \label{tab:permnist_acc}
\end{wraptable}
Just as in Ref.~\cite{synaptic_intelligence}, we do a grid search over the hyperparameters on the first task using a heldout validation set. For the remaining tasks, we settle on learning rate of~$0.002$ and $L^1$ weight regularization ${\alpha=10^{-7}, 10^{-5}, 10^{-6}}$ respectively for the first, second and final layers. Finally, when sparsifying after training each task, we allow for graceful forgetting with a small margin of $m=0.05\%$. We run the experiment 5 times and report the meand and standard deviation of the test accuracy of the network in Tab.~\ref{tab:permnist_acc}.  With test error within $0.05\%$ of single task SGD training, CLNP virtually eliminates catastrophic forgetting in this example and noticeably outperforms previous methods while employing a more restrictive fixed capacity single head architecture.

\setlength{\intextsep}{10pt}
\begin{wrapfigure}{r}{0.4\textwidth}
\vspace{-10pt}
% \fbox{
\includegraphics[clip, trim=0.26cm 0.26cm 0.26cm 0.3cm, width=\textwidth]{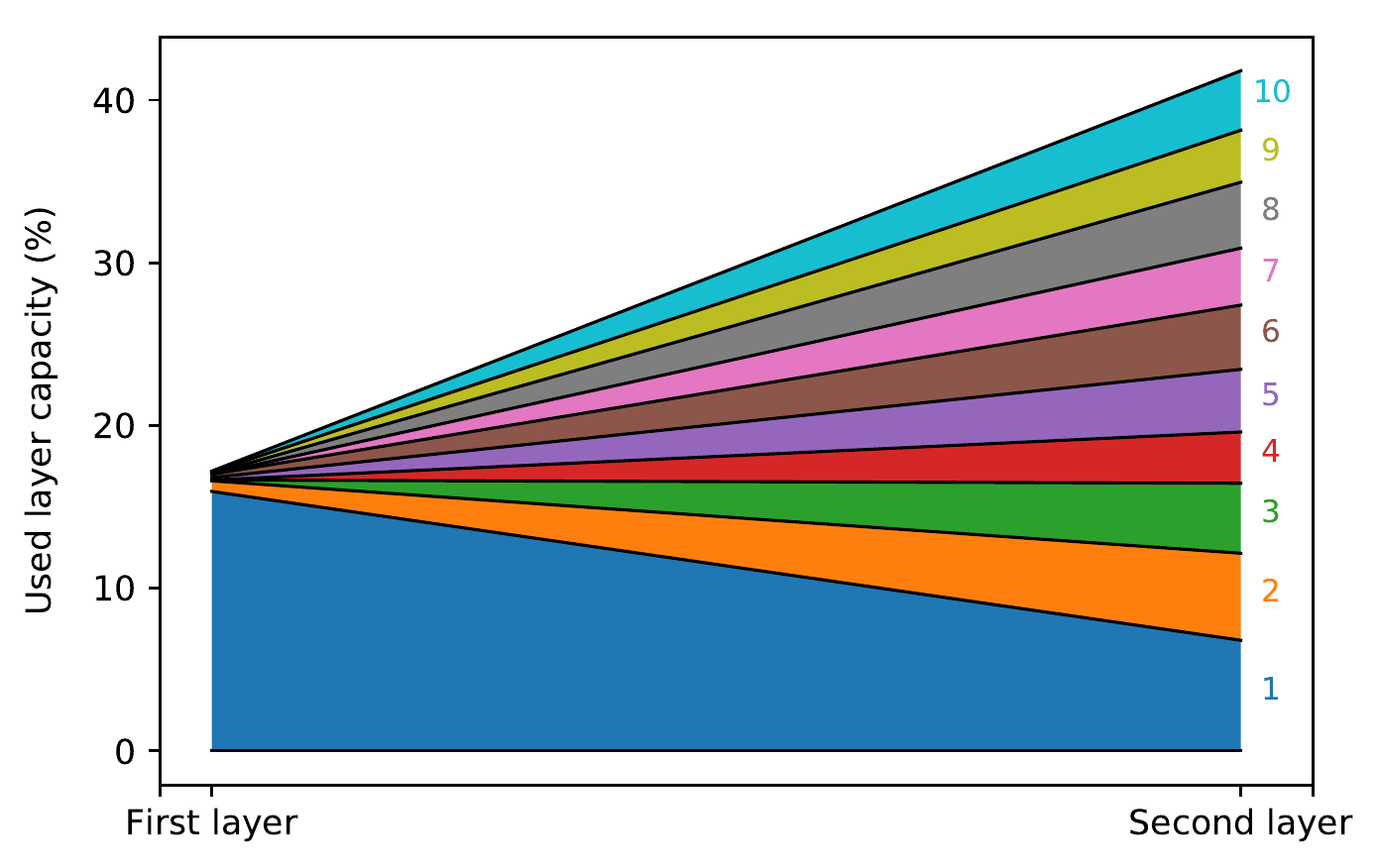}
% }
\caption{\small The percentage of the hidden layer neurons used for each task on the permuted MNIST experiment.} 
\label{fig:permnist_usage}
\end{wrapfigure}

A nice feature of our sparsity based approach is that we have a clear understanding of how much of the network has been taken up by previous tasks and how much is left free. We also have a good indication of how many of the features of the previously learned tasks are being reused.   Fig.~\ref{fig:permnist_usage} shows what percentage of the neurons of the hidden layer are used after each task averaged over the 5 runs. We see that the number of used neurons of the first layer does not greatly increase after the first task is trained, implying a significant amount of reusage of the features of this layer. The number of used neurons in the second layer grows linearly with each task. This is as expected, since in a single-head structure none of the neurons of the final hidden layer are reused in order to prevent interference (see Fig.~\ref{fig:heads}).

\setlength{\intextsep}{12pt}

Finally note that after training all 10 tasks, the neurons of the two hidden layers of the network are only 18\% and 40\% utilized, leaving a lot of capacity free for future tasks. In our experiments, we can train a total of about 25 random permutation tasks with the same test accuracy of 98.4\% before the capacity of the final hidden layer is fully depleted. This is in contrast to previous work where the average accuracy over all tasks continues to decrease as more tasks are trained (see for example Fig.~4 in Ref.~\cite{synaptic_intelligence}).

\subsection{Split CIFAR-10/CIFAR-100} \label{sec:icifar}

In this experiment, we train a model sequentially, first on CIFAR-10 (task 1) and then on CIFAR-100 split into 10 different tasks, each with 10 classes (tasks 2-11). We use two different models for this experiment, a smaller multi-head network used in Ref.~\cite{synaptic_intelligence} and a wider single-head network for demonstration purposes, respectively given in Tab.~\ref{tab:narrow_net} and Tab.~\ref{tab:wide_net}.

\begin{figure}[ht]
\centering
\begin{subfigure}{0.504\textwidth}
\centering
% \fbox{
\includegraphics[clip, trim=0.26cm 0.31cm 0.36cm 0.36cm, width=\textwidth]{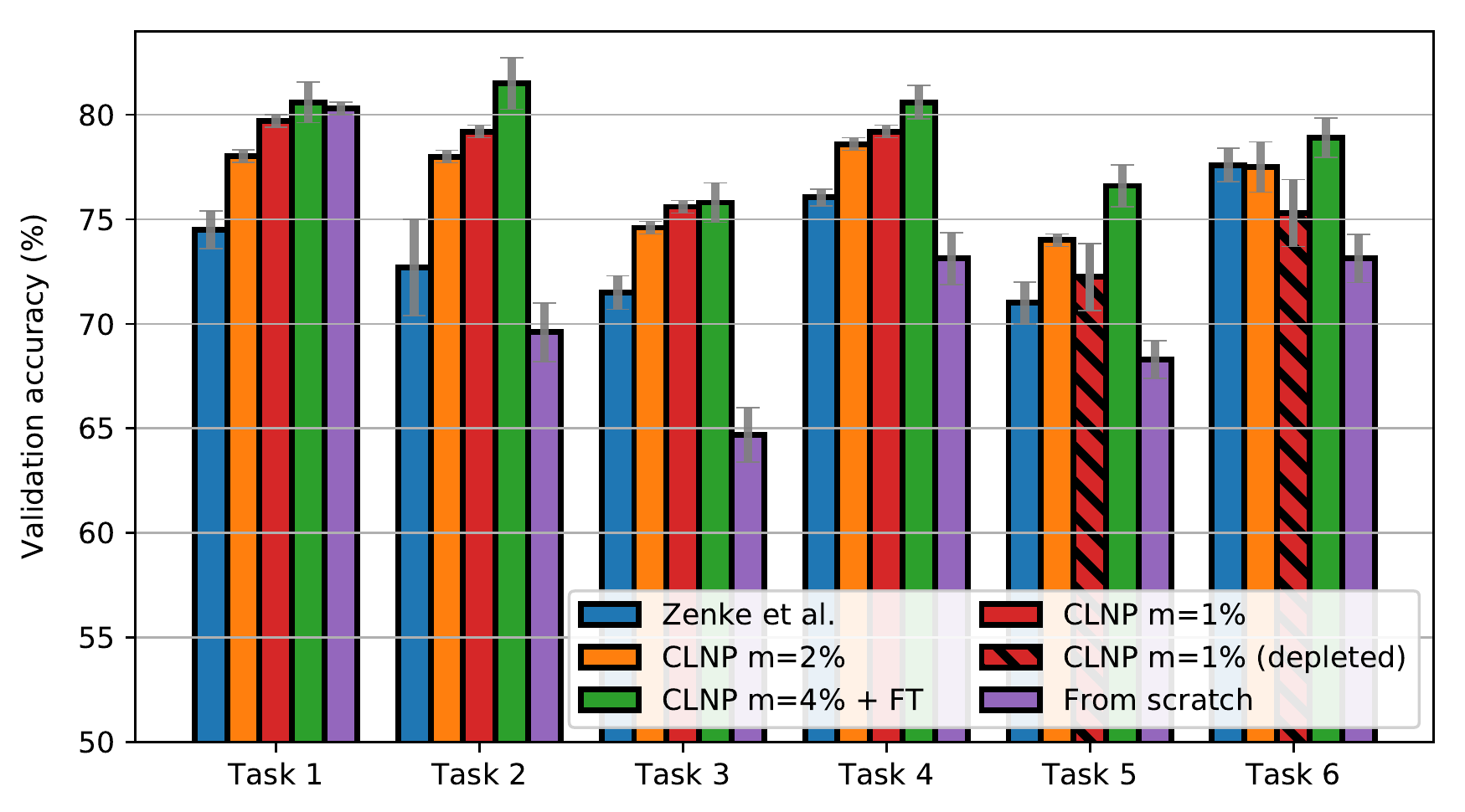}
% }
\caption{\small Validation accuracy comparison}\label{fig:icifar_compare}
\end{subfigure}
\hspace{0.05\textwidth}
\begin{subfigure}{0.396\textwidth} 
\centering
% \fbox{
\includegraphics[clip, trim=0.26cm 0.31cm 0.30cm 0.36cm, width=\textwidth]{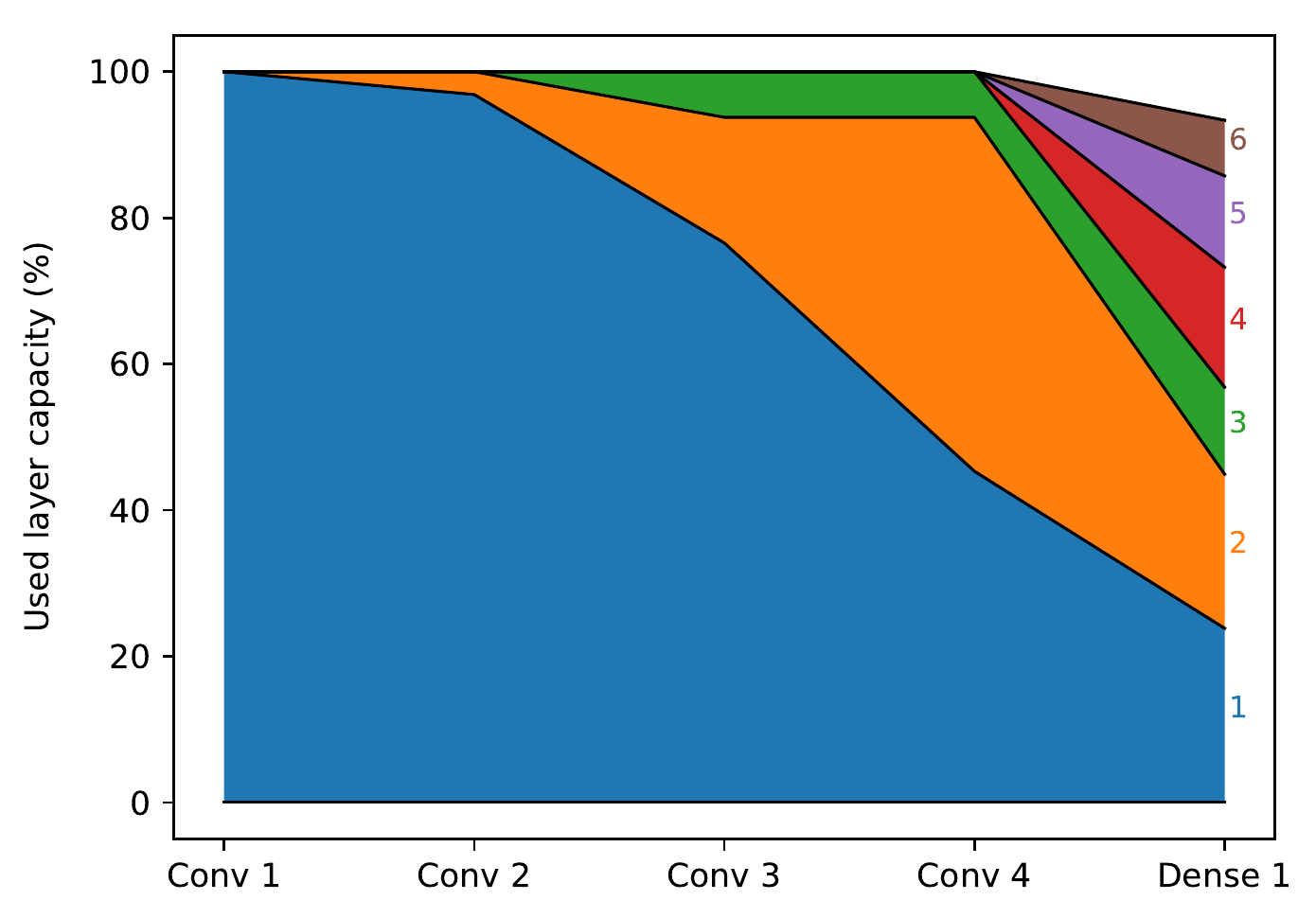}
% }
\caption{\small Average network capacity usage per task}\label{fig:narrow_usage}
\end{subfigure}
\caption{\small CIFAR-10 and split CIFAR-100 results on multi-head network.}\label{fig:narrow_results}
\end{figure}

\setlength{\intextsep}{12pt}
\begin{wraptable}{R}{0.4\textwidth}
  \vspace{-2pt}
  \scalebox{0.65}{
  \begin{tabular}{ccccccc}
    \toprule
    Layer               &  Chan.    &    Ker.        &   Str.     &   Pad.      &   Dropout  \\
    \midrule
    $32\times32$ input  &   3        \\
    Conv 1              &   32      &   $3\times3$    &   1       &      1      &           \\
    Conv 2              &   32      &   $3\times3$    &   1       &      1      &           \\
    MaxPool             &           &   $3\times3$    &   2       &      1      &    0.25   \\
    Conv 3              &   64      &   $3\times3$    &   1       &      1      &           \\
    Conv 4              &   64      &   $3\times3$    &   1       &      1      &           \\
    MaxPool             &           &   $3\times3$    &   2       &      1      &    0.25   \\
    Dense 1             &   512     &                 &           &             &    0.5    \\
    \midrule
    Task 1: Dense       &   10\\
    $\cdots$ : Dense    &   10\\
    Task 6: Dense       &   10\\
    \bottomrule
  \end{tabular}}
  \caption{\small Split CIFAR narrow multi-head model. Convolutional layers and Dense 1 are followed by ReLU activation.}
  \label{tab:narrow_net}
\end{wraptable}
In order to provide a direct comparison to previous results on this dataset we adopt the model and the training scheme of Ref.~\cite{synaptic_intelligence}. Specifically, we train sequentially on only the first 6 tasks of this problem using Adam optimizer with learning rate 0.001. We choose $L^1$ weight regulator coefficient $\alpha=5\times10^{-5}$. We also use two different graceful forgetting schemes defined via validation accuracy acceptability margins of  $m=1\%$ and $m=2\%$. We perform the experiment 5 times and report the validation accuracy and its standard deviation on the 6 tasks. 

\setlength{\intextsep}{12pt}

The results of the experiment are shown in Fig.~\ref{fig:icifar_compare}. We see that again we outperform the previous results by a noticeable margin. Note, however, that the more ambitious $m=1\%$ scheme which only allowed for an initial graceful forgetting of less than 1\%, runs out of capacity after the fourth task is trained. As mentioned in Sec.~\ref{sec:method}, a multi-head network which runs out of capacity on the last hidden layer is trained just like final layer transfer learning, i.e. the previous weights are all fixed and a new head is trained. We notice that after the model capacity is depleted, the performance of the $m=1\%$ scheme plummets, showing the necessity for new neurons to be trained in the core of the network. The more moderate forgetting scheme $m=2\%$, however, maintains high performance throughout all tasks and does not run out of capacity until final task is trained.

Note that even though we outperform previous methods, the narrow structure of this network is ill-suited for our purposes. An ideal network for our method would have an evenly over-parametrized capacity at all layers of the structure but with this network, 95\% of the parameters are concentrated in the dense layer which follows the convolutions. Furthermore, the use of 3 dropout layers is not the most conducive for structural sparsity.  We therefore expect the network to fill up very quickly during training. Fig.~\ref{fig:narrow_usage} shows the network capacity usage per task for the moderate $m=1\%$ forgetting scheme averaged over the 5 runs. Notice that the first task alone almost fills up the entirety of the first and second convolutional layer capacities, leaving little room for new convolutional channels to be learned in future tasks. The fact that even with this undesirable structure we still outperform previous methods is surprising and is an idication that a very large amount of transfer learning is taking place. 

In what follows, we look at two variations on our approach to this problem: first with a slightly more advanced sparsification scheme and second with a single-head network of much wider width.

\paragraph{CLNP + fine tuning.} In sparsification literature, following the pruning of the redundant weights and neurons, it is common practice to fine-tune the remaining weights of the network and iterate on this process until a desired sparsity is achieved. In this paper we have chosen to perform a single iteration of this process and also omit the fine-tuning stage for the sake of simplicity and also so that our results are directly comparable to previous work. We now look at the potential of compression based continual learning approaches given a slightly more advanced variation of our sparsification scheme. Explicitly, after training on any task, we still perform a single iteration of neuron pruning but this time with a larger graceful forgetting margin of $m$ = 4\% and then we fine-tune the remaining weights of the network by retraining on the same task for 20 epochs with learning rate $\xi=10^{-4}$ and $L^1$ weight regulator $\alpha=5\times10^{-5}$. The results of this method are given in Fig.~\ref{fig:icifar_compare} under `CLNP $m$ = 4\% + FT'. We see that here there is virtually no catastrophic forgetting on the first task (if anything the model performs even better after pruning and retraining as has been reported in previous sparsity literature \cite{brain_damage,trimming}). The remaining tasks also get a significant boost from this improved sparsification method.

\begin{wraptable}{R}{0.4\textwidth}
  \vspace{-8pt}
  \scalebox{0.65}{
  \begin{tabular}{ccccccc}
    \toprule
    Layer               &  Chan.    &    Ker.        &   Str.     &   Pad.      &     \\
    \midrule
    $32\times32$ input &   3        \\
    Conv 1             &   128      &   $3\times3$    &   1       &      1      &           \\
    % Batch Norm         &            &                 &           &             &   ReLU    \\            
    Conv 2             &   256      &   $3\times3$    &   2       &      0      &           \\
    % Batch Norm         &            &                 &           &             &   ReLU    \\
    Conv 3             &   512      &   $3\times3$    &   1       &      0      &           \\
    % Batch Norm         &            &                 &           &             &   ReLU    \\
    Conv 4             &   1024     &   $3\times3$    &   2       &      1      &           \\
    % Batch Norm         &            &                 &           &             &   ReLU    \\
    Conv 5             &   2048     &   $3\times3$    &   1       &      0      &           \\
    % Batch Norm         &            &                 &           &             &   ReLU    \\
    Conv 6             &   10       &   $3\times3$    &   1       &      0      &           \\
    AvgPool            
    \\    
    \bottomrule
  \end{tabular}}
  \caption{\small Split CIFAR wide single-head model. Convolutions 1-5 are followed by BatchNorm then ReLU activation. }
  \label{tab:wide_net}
\end{wraptable}

\paragraph{Wide single-head network.} To gain further insight into the interplay of transfer learning and network width in networks with activation based neuronal sparsity, we also trained a much wider fully convolutional network on the same task (Tab.~\ref{tab:wide_net}). This time we use a single-head structure making the task more challenging because of the fixed total capacity of the network. We also use a slightly different training scheme, employing a heldout validation set comprised of 10\% of the training samples to find optimal hyperparameters for each task and for early stopping. We run the experiment 5 times and report the mean and standard deviation of the test accuracy after all tasks have been trained. 

\begin{figure}[ht]
\centering
\begin{subfigure}{0.504\textwidth}
\centering
% \fbox{
\includegraphics[clip, trim=0.26cm 0.31cm 0.36cm 0.36cm, width=\textwidth]{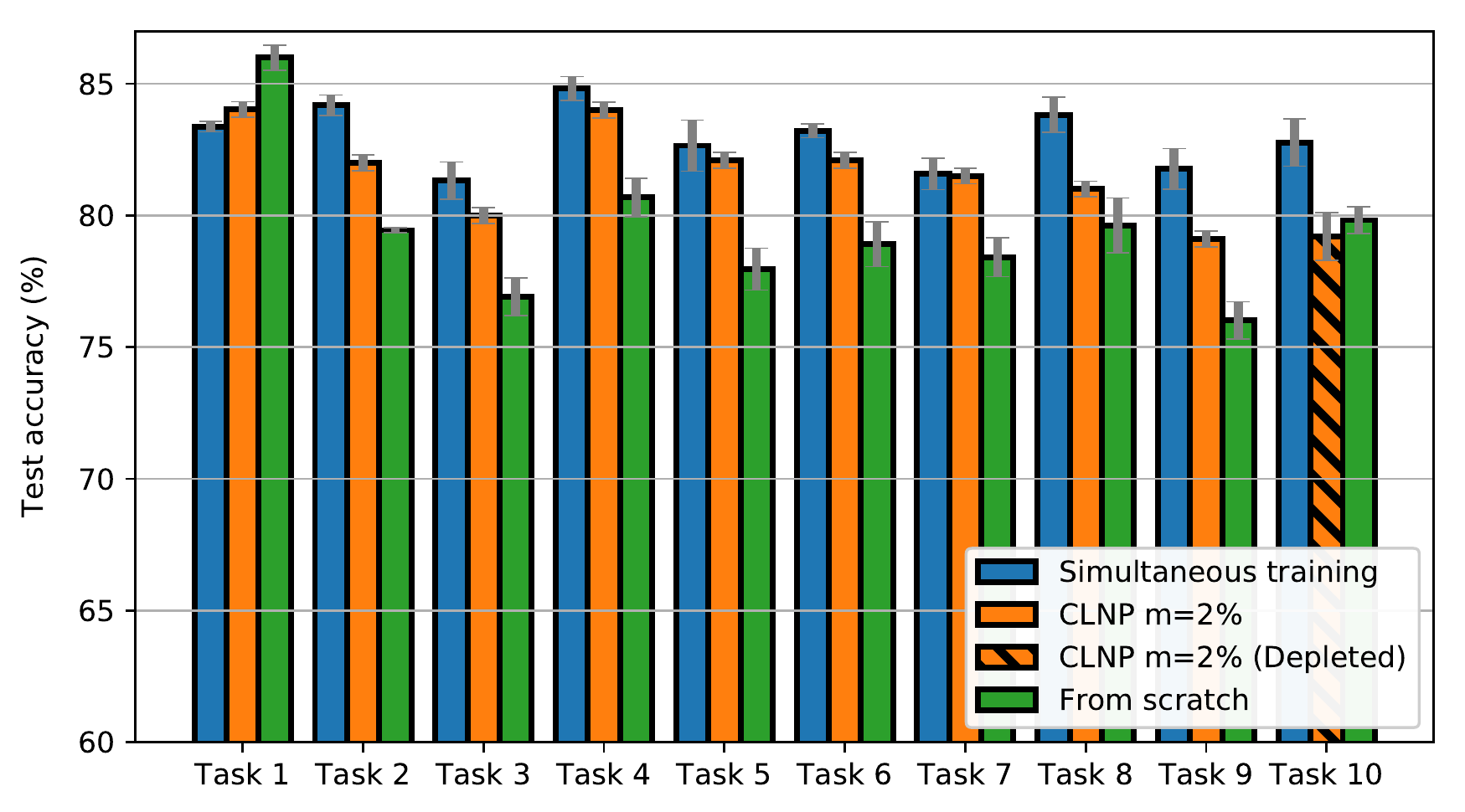}
% }
\caption{\small  Test accuracy comparison}\label{fig:wide_accuracy}
\end{subfigure}
\hspace{0.05\textwidth}
\begin{subfigure}{0.396\textwidth} 
\centering
% \fbox{
\includegraphics[clip, trim=0.26cm 0.31cm 0.33cm 0.36cm, width=\textwidth]{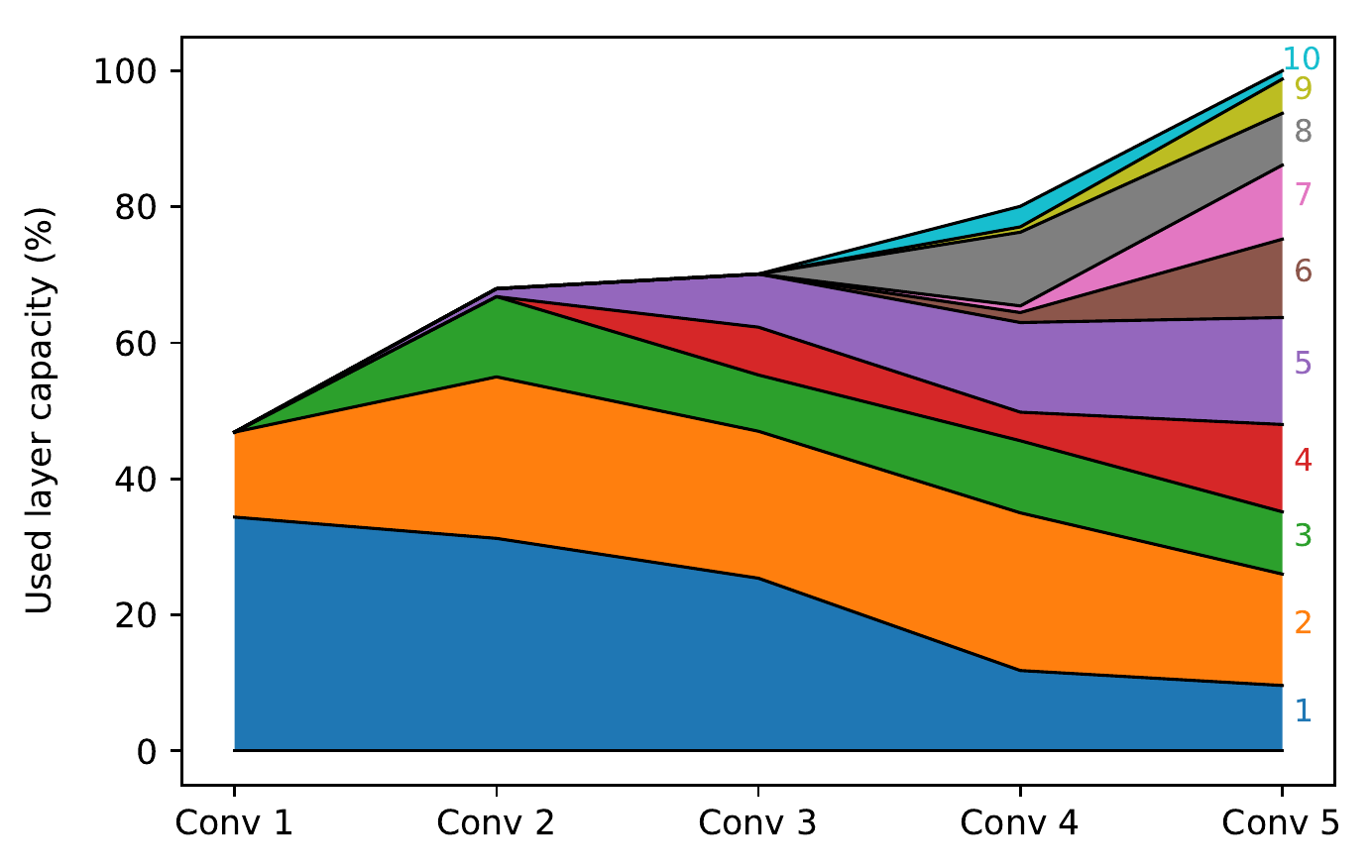}
% }
\caption{\small Sample network capacity usage per task}\label{fig:wide_usage}
\end{subfigure}
\caption{\small CIFAR-10 and split CIFAR-100 results on wide single-head network.}\label{fig:wide_results}
\end{figure}

The wider network in this experiment has the capacity to train 10 tasks (i.e. CIFAR-10 plus 9 out of the 10 CIFAR-100 tasks) before depleting its final hidden layer. Unlike multi-head structures where a depleted final hidden layer leads to naive transfer learning on new tasks, for single-head structures, a depleted final hidden layer simply has no free connections to the output and the network is entirely fixed. Fig.~\ref{fig:wide_accuracy} shows the results of this experiment.\footnote{The only comparable single-head experiment on this dataset that we are aware of was done by~\citet{superposition} who also use a single-head 6-layer convolutional network. Using superposition techniques they find that the first task suffers about a 10\% catastrophic forgetting (from $\sim72\%$ down to $\sim63\%$ after training 4 new tasks, accuracy on other tasks not reported). In comparison our method suffers less than a 2\% drop via graceful forgetting and remains unchanged during the 9 subsequent tasks.} For comparison we have also provid
ed the results of multi-task training and training each task individually from scratch. In order to not bias our choices of graceful forgetting margins, the multi-task and individual training computations were performed only after CLNP results were completed. Details of the multi-task training is given in the supplementary materials section.

The average neuron capacity usage per task is given by Fig.~\ref{fig:wide_usage}. There are a number of interesting features in this graph. First note that similar to the single-head MNIST usage graph, tasks number~1-7 each  roughly takes up the same number of neurons of the final hidden layer. This is again as expected since the neurons of this layer are connected to the output layer and are not reused in order to avoid interference. By the time we get to task 8, more than 85\% of the capacity of this layer is depleted. This directly affects the performance of the network which can be seen in terms of reduced test accuracy both in comparison to the results from multi-task training and individual training from scratch. 

Perhaps the most interesting observation in the training of the wide network is in the number of new channels learned  at each layer for each consecutive tasks. Notice that the first convolutional layer trains new channels only for task 1 and 2. The second and third convolutional layers, grow new channels up to task 3 and task 5 respectively. The fourth layer keeps training new channels up to the last task. The fact that the first layer grows no new channels after the second task implies that the features learned during the training of the first two tasks are fully utilized and deemed sufficient for the training of the subsequent tasks. The fact that this sufficiency happens after training more tasks for layers 2 and 3 is a verification of the fact that features learned in lower layers are more general and thus more transferable in comparison with the features of the higher layers which are known to specialize ~\cite{transferability}. This observation implies that models which hope to be effective at continual learning need to be wider in the higher layers to accommodate this lack of transferability of the features at these scales.

%%%%%%%%%%%%%%%%%%%%%%%%%%%%%%%%%%%%%%%%%%%%%%%%%%%%%%%%%%%%%%%%%%%%%%%%%%%%%%%%%%
\section{Conclusion} \label{sec:conclusion}
%%%%%%%%%%%%%%%%%%%%%%%%%%%%%%%%%%%%%%%%%%%%%%%%%%%%%%%%%%%%%%%%%%%%%%%%%%%%%%%%%%

In this work we have introduced a simple and intuitive lifelong learning method which leverages the over-paremetrization of neural networks to train new tasks in the inactive neurons/filters of the network without suffering any catastrophic forgetting in the previously trained tasks. We also implemented a controlled way of graceful forgetting by sacrificing a little bit of accuracy at the end of the training of each task in order to regain network capacity for training new tasks. We showed empirically that this method leads to noticeably improved results compared to previous approaches. Our methodology also comes with simple diagnostics regarding the number of free neurons left for the training of new tasks. Model capacity usage graphs are also informative regarding the transferability and sufficiency of the features of different layers. Using such graphs, we can verify the notion that the features learned in earlier layers are more transferable. We can also leverage these diagnostic tools to pinpoint  any layers that run out of capacity prematurely, and resolve thse bottlenecks in the network by simply increasing the number of neurons in these layers when moving on to the next task. In this way, our method can efficiently expand to accomodate more tasks and compensate for sub-optimal network width choices. 

It is important to note that our algorithm is crucially dependent on the sparsification method used. In this work we employed a neuron activity sparsification scheme based on a single iteration neural pruning. We also demonstrated that using a slightly more advanced sparsification scheme, i.e. adding a fine tuning  step after pruning can lead to even better results. This shows that our results, regarding the number of tasks that can be learned or the accuracy that can be kept in a fixed network structure, can only improve with more efficient sparsification methods. We hope that this observation opens the path for a new class of neuronal sparsity focused lifelong learning methods as an alternative to current weight elasticity based approaches.

\subsubsection*{Acknowledgments}

We would like to thank Kyle Cranmer and Johann Brehmer for interesting discussions and input. SG is supported by the James Arthur Postdoctoral Fellowship. MK is supported by the US Department of Energy (DOE) under grant DE-AC02-76SF00515 and by the SLAC Panofsky Fellowship. This work was partly supported by NVidia (Project: "NVIDIA - NYU Autonomous Driving Collaboration").

\appendix

\section{Multi-task training on CIFAR-10/100}

Multi-task learning, the simultaneous training of multiple tasks at the same time, is generally a complicated problem. Extra care has to be taken especially when the different tasks have different difficulties or different dataset sizes. In the mixed CIFAR-10/100 problem, the dataset for the first task is ten times larger than the other tasks.  An important practical choice which affects the relative performance of the tasks is how to perform minibatch training. In particular during one epoch of training, if we draw from all tasks indiscriminately, the samples of the smaller datasets run out when only 10\% of the larger dataset is seen. In this case we can either start a new epoch i.e. cut the larger dataset short or we can keep training on the larger dataset until it runs out. In practice, this choice amounts to either sacrificing the performance of the first task in favor of the other tasks or vice versa. 

For the purposes of comparison to lifelong learning, we take a compromise approach. When training all tasks at the same time, we take each batch to comprise of 50 samples from task 1 and 10 samples from each of the other tasks, the total batch size adding to 140. We then start a new epoch when the the smaller datasets run out. In this way, a new epoch is started when only 50\% of the large dataset of task 1 is seen. Note that the relative number of samples in each task biases the network towards one task or another, therefore picking an even higher number of samples of the first task would again lead to sacrificing the accuracy on the other tasks. We made the choice of the 5 to 1 ratio purely as a middle ground. It is possible that other choices can lead to better overall performance.

In order to adapt the single-head model of Tab.~\ref{tab:wide_net} for multi-task training, we partition the neurons of the final hidden layer into 10 equal parts which form the ``heads'' of the 10 different tasks. We chose to train only on 10 out of the 11 tasks in order to provide a fair comparison since our continuous learning algorithm depletes the network after 10 tasks. We train using 120 epochs using Adam optimizer with learning rate 0.001 and learning rate schedule with milestones at 50 and 90 epochs and $\gamma=0.05$. We use the same heldout validation set as continuous learning and individual learning for early stopping. We run the training 5 times and report the mean and standard deviation of the test accuracy in Fig.~\ref{fig:wide_accuracy}.

\bibliographystyle{ACM-Reference-Format}
\bibliography{unforgettable}

\end{document}